\documentclass[11pt]{article}

\usepackage[preprint]{acl}

\usepackage{times}
\usepackage{latexsym}

\usepackage{booktabs}
\usepackage{amsmath}
\usepackage{float}
\usepackage{booktabs}
\usepackage{tabularx}
\usepackage{array}
\usepackage{multirow}

\usepackage[T1]{fontenc}

\usepackage[utf8]{inputenc}

\usepackage{microtype}

\usepackage{inconsolata}

\usepackage{graphicx}

\title{Detecting Data Contamination in Large Language Models}

\author{Juliusz Janicki \\ University of Amsterdam  \\ jjanicki111@gmail.com\And
        Savvas Chamezopoulos \\ Elsevier  \\ s.chamezopoulos@elsevier.com 
        \AND
        Evangelos Kanoulas \\ University of Amsterdam  \\ E.Kanoulas@uva.nl \And
        Georgios Tsatsaronis \\ Elsevier  \\ g.tsatsaronis@elsevier.com}

\begin{document}
\maketitle
\begin{abstract}
Large Language Models (LLMs) utilize large amounts of data for their training, some of which may come from copyrighted sources. Membership Inference Attacks (MIA) aim to detect those documents and whether they have been included in the training corpora of the LLMs. The black-box MIAs require a significant amount of data manipulation; therefore, their comparison is often challenging. We study state-of-the-art (SOTA) MIAs under the black-box assumptions and compare them to each other using a unified set of datasets to determine if any of them can reliably detect membership under SOTA LLMs. In addition, a new method, called the Familiarity Ranking, was developed to showcase a possible approach to black-box MIAs, thereby giving LLMs more freedom in their expression to understand their reasoning better. The results indicate that none of the methods are capable of reliably detecting membership in LLMs, as shown by an AUC-ROC of approximately 0.5 for all methods across several LLMs. The higher TPR and FPR for more advanced LLMs indicate higher reasoning and generalizing capabilities, showcasing the difficulty of detecting membership in LLMs using black-box MIAs. 

\end{abstract}

\section{Introduction}

Large Language Models (LLMs) have seen substantial performance gains in recent years \cite{LLMsurvey}. Most of the improvement in performance comes primarily from training LLMs on progressively larger datasets \cite{EducatingSilicon2024}. However, the large amount of data used for training LLMs makes it impossible to exclude all unwanted documents. Those documents might contain parts of benchmark datasets \cite{Oren2024}, Personally Identifiable Information (PII) \cite{Nakka2024}, or copyrighted content like novels \cite{Chang2023}. The inclusion of these documents can result in serious privacy concerns and potentially lead to lawsuits against the companies developing the LLMs \cite{NYT2023}. Due to these reasons, it is important to have robust and accurate methods for detecting content used in the training of LLMs. Membership Inference Attacks (MIA) \cite{Hartmann2023} are a type of privacy attack that aims to give concrete evidence that a data point is a member or non-member of a training corpus. MIAs are widely used for standard machine learning (ML) approaches and have recently gained popularity in the context of LLMs.

The existing methods for MIA range from white-box methods, which use the model weights \cite{Stoehr2024}, to a more popular family of gray-box methods, which use the logits and probabilities of the output tokens \cite{Xie2024[RECALL], Zhang2024_Min-kpp, Shi2024}. The third family of MIA methods views LLMs as a black-box, hence their name "black-box methods" \cite{Chang2023}. The black-box methods are the most preferable as they can be applied to any LLM, as they only require the plain output of the model. However, as black-box methods do not have any access to the internal states of the LLM, they require several assumptions \cite{Fu2024} and, although not exclusive to black-box methods, are difficult to evaluate due to the lack of ground truth labels \cite{Zhang2024[FPR_def]} and the lack of knowledge of training data. Therefore, MIA black-box methods have become less popular compared to the gray-box method family, which yields almost perfect results on benchmark datasets \cite{Kim2024}.

Black-box methods are generally built around two key components: a “hook,” placed in the prompt, that guides the LLM, and a modification in the data to test whether the large language model has memorized the text. The most common “hook” is a title of the article, paper, or book that guides and narrows the scope of the answers that the LLM can provide \cite{Duarte2024, Karamolegkou2023}. Another hook can be a true prefix showing the large language model the type of answer one is trying to obtain \cite{Nakka2024}. The second component of most black-box methods, the data manipulation, is the part that helps distinguish whether the large language model has memorized a piece of text or not. The original pieces of text often serve as ground truth labels, and a modification of the original text plays the role of a task that the model must perform, given a prompt. If the match between the original and modified text is correct, one can argue that the large language model has memorized the text. If not, then it can be labeled as a non-member data point, transforming the task into a binary classification. Due to the dependency of these methods on data manipulation, black-box methods are often difficult to compare with one another, as obtaining a unified dataset can prove challenging \cite{Duarte2024, Karamolegkou2023, Nakka2024, Chang2023}.

This paper aims to evaluate current state-of-the-art methods under a unified set of datasets to identify member data in popular LLMs in a black-box setting, and to test a new method that further expands our understanding of membership inference attacks on LLMs. In particular, we investigate whether there are reliable black-box methods that can present significant evidence of the inclusion of data in the training corpus of an LLM. Alongside this main objective, we examine to what extent a familiarity scale, applied as a black-box method, can reliably detect training data contamination in large language models, whether large language models are becoming too capable at reasoning to distinguish between memorization and reasoning, and whether recent large language models implement safeguards to prevent the reproduction of verbatim or copyrighted material.

The proposed research makes the following contributions to the scientific field of Membership Inference Attacks in LLMs:

\begin{enumerate}
\item We propose a novel method for detecting data contamination in the black-box setting (only using the output) of large language models (LLMs).
\item We evaluate state-of-the-art membership inference methods on the latest black-box large language models, which have not been previously assessed in the academic literature.
\item We present a comparison of black-box methods on a unified dataset.
\end{enumerate}

\section{Related Work}
\label{sec:related_work}
In this section, literature related to MIA for LLMs will be explored. The papers in each subsection will be discussed in chronological order starting from the oldest ones.

\subsection{Gray-box Methods}
The gray-box model family is the most popular for MIA. Gray-box methods entail access to the logits/probabilities of the output tokens on the LLM. Not all of the models provide access to the logits; therefore, gray-box methods can only be applied to some. One of the first gray-box methods, called Min-k\%, used the hypothesis that a non-member data point should contain a few tokens with low probabilities. In contrast, a seen example would contain fewer of those tokens with low probabilities \cite{Shi2024}. Min-k\% inspired the method known as Min-k\%++, which builds upon the Min-k\% method \cite{Zhang2024_Min-kpp}. The Min-k\%++ methods suggested that the training data should be the local maximum along the input dimension, due to maximum likelihood training. This means that if one token, which is similar to other tokens, stands out by having a higher probability of being generated, it is likely to be memorized. Both of these approaches have become standard baselines in examining MIA in LLMs from the token probabilities standpoint; however, they fall short of the newer methods. 

Another gray-box approach to examining MIA in LLMs has emerged due to the hypothesis that prefixing an input with a non-member prefix results in a larger shift in distribution for member data than for non-member data points \cite{Xie2024[RECALL]}. The original method, called Relative Conditional Log-Likelihood (RECALL), applied this hypothesis by dividing the log-likelihood of the input given a non-member prefix by the log-likelihood of just the input \cite{Xie2024[RECALL]}. This approach, however, was later proven to be faulty as RECALL in the original paper used a specifically selected prefix which boosted its performance \cite{Kim2024}. Nonetheless, the idea of prefixing the input is still valid, and has been improved upon \cite{Wang2024[CON-Recall]}. CON-RECALL leverages the idea that member prefixes boost the log-likelihood for member data but reduce it for non-member data, and vice versa \cite{Wang2024[CON-Recall]}. CON-RECALL also uses a regularization term to enhance the difference between a data point given a member and non-member prefix \cite{Wang2024[CON-Recall]}. The newest methodology, which incorporates this concept, is a general framework for fitting the best prefix to a given input using expectation maximization, referred to as EM-MIA \cite{Kim2024}. EM-MIA repetitively updates the prefix score given the membership score by selecting different prefixes to obtain the best likelihood of a data point being a member. EM-MIA fixes the problem in the RECALL paper by using the expectation maximization to find the best prefix instead of giving the best prefix, which is not always known. EM-MIA gets almost perfect results in the task of MIA for LLM, in the gray-box setting \cite{Kim2024}.

The last gray-box approach for MIA in LLMs, which is unique, is based on the human emotion of being surprised \cite{Zhang2024[SURP]}. The idea behind this method, called SURP, is that if an LLM is sure about an output (has low Shannon entropy) and has a low probability of the golden label, it is unlikely that it has memorized the text. Translating it to the idea of being surprised, the LLM is 'surprised' that the golden label is what it is and not the one that it predicted. SURP obtains better results than the Min-k\% approaches but falls short of the RECALL approaches \cite{Zhang2024[SURP]}.

\subsection{Black-box Methods}
The black-box methods are the most desired types of methods, as they can be applied to any LLM, as the only input for the methods is the plain text that the LLM outputs. The first significant attempt used Name Cloze Queries \cite{Chang2023}. The idea behind this method was to mask a named entity in the input and have the LLM guess the masked word. If the LLM was able to guess the input, it was hypothesized that it had memorized the text. This method, however, has some significant flaws, such as not taking into consideration the model's generalization capabilities or hallucinations, which could be used to obtain the missing word. A more robust solution is suggested to make the LLM output a whole chunk of text given a prefix of that chunk \cite{Karamolegkou2023}, rather than just making the LLM guess one word. Another solution suggested giving the LLM a multiple-choice question and masking one of the incorrect answers, making the LLM fill in the masked answer and provide the correct answer to the multiple-choice question \cite{Deng2023}. To diminish the prevalence of hallucinations and generalization capabilities of LLMs, one proposed method involves presenting the LLM with a multiple-choice question in which all incorrect options are paraphrased versions of the correct answer \cite{Duarte2024}. Other methods have also attempted to use prompt engineering to trick the LLM into outputting PII \cite{Nakka2024} or to utilize the edit distance between the LLM's output and the truth labels as an indicator of memorization \cite{Dong2024}. Two other methods \cite{Oren2024, Song2024} have been based on the property of datasets called exchangeability, where they tested whether the model prefers the canonical ordering of the data. Lastly, LiMem \cite{Xie2024[LiMem]} is based on the idea that memorization is part of generalization, and just like students who might memorize concepts but struggle when the question about the idea is slightly modified, the same might happen with LLMs. Based on this, the methods present reasoning problems to the LLM from popular websites, along with the same problem with a slightly modified question, to determine if the LLM can answer both correctly. 

\subsection{Challenges in MIA for LLMs}
The problem of membership inference attacks for LLMs is defined as follows: given an input sequence x = $x_1$, $x_2$, ..., $x_n$ and a pre-trained model M, infer if x was a part of the dataset D used for training M \cite{Zhang2024[SURP]}. MIA for LLMs presents several challenges, the first being that it is difficult to distinguish what memorization entails, as there are many types of memorization beyond verbatim text. For example, two other types of memorization, besides verbatim memorization, that an LLM can present are factual memorization, which involves recalling particular information from the training data of the LLM based on context cues, or conceptual memorization, which involves the models generalization skills to reproduce information based on ideas and concepts from its training data\cite{Satvaty2024}. 

Other challenges concern the estimation of memorization, as the retrieved information by the large language models can be due to hallucinations or the reasoning and generalizing skills of the LLM \cite{Hartmann2023}. Hallucination refers to the LLM generating content that seems reasonable for the given prompt yet incorrect, which can come from many stages, like training or inference, or even the data that the LLM was trained on \cite{Hallucination}. The generalizing and reasoning skills of LLM become better with each new model \cite{Artificial_Analysis}, making it more difficult to distinguish whether the model can retain the piece of text from memorization or if it was reasoned based on the context of the given passage \cite{Hartmann2023}.

Another challenge is the lack of gold truth labels, as the training data for LLMs is often not disclosed. One can only assume that content from widely used sources before the LLM's cut-off date was likely included in the training set, while data published after the cut-off date was not. Therefore, data before the cut-off date of the LLM is often considered to be member data, and data after the cut-off date is considered non-member data. Due to these challenges, there are a lot of assumptions around each of the methods \cite{Fu2024}. The assumptions of the other methods will be presented in Section 3.

\subsection{Research gap}
Recent studies on membership inference attacks often focus on the same large language models, despite the availability of newer versions. Popular LLMs commonly evaluated include early models such as Claude and the Llama family \cite{Karamolegkou2023, Deng2023, Duarte2024}. This paper aims to explore how more recent models respond to existing MIA techniques and whether their outputs are consistent with those presented in previous papers. Additionally, black-box MIA studies rarely compare their methods against other black-box approaches, primarily due to the challenge of acquiring data that is satisfactory for all methods or the difficulty in devising a unified comparison method. This work aims to address this gap by directly comparing several black-box MIA techniques over a unified set of datasets.

\section{Methodology}
\label{sec:methodology}
The methodology for this paper consists of two main parts. The first part involves constructing a pipeline to evaluate state-of-the-art (SOTA) methods across two datasets. The second component focuses on designing and implementing a novel approach for membership inference attacks, assuming black-box access to the large language model.

\begin{figure*}[t]
\noindent\fbox{%
  \parbox{\textwidth}{%
    \hspace*{2em}Please rank the following three text snippets based on how likely they are to appear in the well-known paper titled: \{title\}.\\
    \hspace*{2em}Order them from most likely (1) to least likely (3). Note that the passages are not presented in any particular order.\\
    \\ 
    \hspace*{2em}Text snippets:\\
    \hspace*{2em}\{chunks\}\\
    \\ 
    \hspace*{2em}Respond only with a list of integers, for example: 3, 1, 2
  }%
}
\caption{Prompt used for Familiarity Ranking}
\label{fig:FR}
\end{figure*}

\subsection{Evaluation of SOTA methods}
\subsubsection{SOTA methods}
As discussed in the related works, there are numerous MIA methods for LLMs. If time were not a constraint, comparing all of them over the datasets would be ideal. However, due to the time constraint, only a few of them can be used as baselines. The primary methods we implemented are: Name Cloze Queries \cite{Chang2023}, DE-COP \cite{Duarte2024}, and Probing \cite{Karamolegkou2023}. These are all Black Box Methods relying solely on the output of the LLM.

The first method, Name Cloze Queries, requires one of the words from the input to be masked. The user then asks the LLM to predict that masked word. The masked word is a proper entity or name, as generating it is least likely to be reasoned or generated, compared to, for example, a preposition or verb. It can be formulated in the following way: 

\begin{equation}
\mathrm{NCQ}(x_{\text{mask}}) = w \mid x_{\text{mask}}
\end{equation}

where $\boldsymbol{x}$ is the text and $\boldsymbol{w}$ is the word.

Name Cloze Queries work under the assumption that an exact match of the word under the [MASK] token indicates that the LLM has seen this passage in its training data. Therefore, labeling an exact match as a member and a mismatch as a non-member. 

The DE-COP method prompts the LLM with a multiple-choice question about which one of the passages is from a given source, for example, arXiv. One of the answers is the correct passage, and the other three are paraphrased versions of that passage. The model's task is to choose the correct one. DE-COP can be expressed as: 

\begin{equation}
p_{\text{selected}} = \arg\max_{p_i} \; \mathcal{M}(p_i \mid q)
\end{equation}

where $\mathcal{M}(p_i \mid q)$ is the likelihood assigned by the language model to passage $p_i$ being the correct response to query $q$. DE-COP works under the assumption that if an LLM correctly chooses the correct answer to the multiple-choice question, it has memorized those pieces of text, and the data is labeled as a member. Otherwise, if the LLM chooses the wrong answer to the multiple-choice question, it is labeled as a non-member. 

The final approach, Prefix Probing, involves prompting the LLM with the beginning of a passage and asking it to generate the suffix, using only the title of the text as context. In the original method, two types of Probing were used: Direct Probing for black-box models, which involved prompting the model to generate the first page of a book based on its title, and Prefix Probing for open-source models, which provided the initial part of the text for continuation. In our case, however, we will use Prefix Probing with black box models. This switch is due to the nature of our dataset. While the original method dealt with books, where generating a first page from a title made sense, we are working with scientific papers and Wikipedia articles, where such a prompt is less meaningful. From now on in the paper, Probing will refer to Prefix Probing. Probing can be mathematically written as:

\begin{equation}
\hat{S} = \arg\max_{S_i} \; \mathcal{M}(S_i \mid T, P)
\end{equation}

In this equation, $S_i$ represents a suffix, and $\mathcal{M}$ denotes the language model. The model generates the suffix $\hat{S}$ given a prefix $P$ and the title of the chunk of text $T$. To determine if the LLM has memorized the text verbatim, a longest common subsequence algorithm is applied between the model's output and the original suffix. If the similarity is $x$ or higher, the output is considered a member; otherwise, it is not. This method tests the verbatim memorization of the LLM, working under the assumption that if a model can infer the suffix, it has seen it in its training data.

\subsubsection{Datasets}
The initial dataset was sourced from the RealTimeData collection \cite{LatestEval}, which is accessible on Hugging Face\footnote{\url{https://huggingface.co/RealTimeData}}. This dataset includes monthly batches of arXiv articles starting from January 2017. For our study, we gathered data from November 2024 to April 2025 as non-members, along with additional data from late 2020, which was obtained as members' data points. These dates were selected based on LLMs' cut-off dates to stay within the models’ training distribution. We decided not to use benchmark datasets like WikiMIA \cite{Shi2024} or MIMIR \cite{Duan2024} because their non-member data points begin in 2023. Since we are testing large language models, some of which were released in late 2024, these non-member data points from WikiMIA and MIMIR might have been part of their training data, making them unsuitable for our evaluation.

From the arXiv papers, which are scrapped in LaTeX, we extracted text chunks of approximately 500 characters, ensuring each included proper named entities to maintain semantic meaning. We intentionally avoided common sections such as the introduction or related work that often contain generalized or well-known information. Instead, we focused on extracting information from more unique sections, such as results, methodology, and discussion, which are more likely to contain novel content specific to each paper. This strategy aims to produce unique text segments that are less likely to be generalized by an LLM and more indicative of memorization. The dataset contains approximately 100 member and 100 non-member data points, where each data point originates from a distinct article. This approach was taken to ensure broad coverage of various articles and to maintain fairness, so that if an LLM has memorized one of the articles, it doesn’t give multiple correct predictions for the content of the same member paper. 

The second evaluation dataset, also obtained from the RealTimeData collection \cite{LatestEval}, is composed of Wikipedia articles. The Wikipedia dataset is organized differently from the arXiv dataset, as identical articles are scrapped from Wikipedia every month. We intend to use this setup to examine whether LLMs are more effective at identifying older versions of the same content. In this case, the assumed member versions of the articles are from December 2020, while the non-member versions are from April 2025. Chunks of approximately 500 characters were extracted from the Wikipedia articles, ensuring each chunk contains at least one proper entity or name to be considered meaningful. The dataset consists of approximately 200 member and 200 non-member data points, representing different versions of the same articles. As a result, around 200 unique articles are included. 

It is important to note that, although sources like Wikipedia and arXiv are commonly included in the training data of LLMs, we cannot be certain that the specific examples labeled as "member" in our datasets were part of the training data; nevertheless, our study proceeds under the assumption that they were.

\subsubsection{Models}
We evaluated the methods on six large language models spanning four different model families: OpenAI \cite{OpenAIModels}, Anthropic \cite{AnthropicModels}, Meta \cite{MetaModels}, and Mistral \cite{mistralmodels}. The specific models used in our experiments are GPT-3.5 Turbo \cite{gpt35turbo_table}, GPT-4o \cite{Gpt4}, GPT-4o Mini \cite{gpt4mini}, LLaMA 3.1 70b \cite{lama31_table}, Mixtral 8x7b \cite{Mixtral}, and Claude 3.5 Sonnet \cite{Claude35Sonnet}. The cut-off dates for all these models are listed in Table~\ref{tab:model_cutoffs}. The cut-off date for Mixtral 8x7B has not been officially provided, but since it was released in January 2024, the cut-off date is likely to be sometime before that. 


\begin{table*}[h]
  \centering
  \begin{tabular}{llll}
    \hline
    Model & Release & Cutoff & Provider \\
    \hline
    GPT-4o & 2024.05 \cite{gpt4o_table} & 2023.10 \cite{open_ai_table} & OpenAI \\
    GPT-4o-mini & 2024.07 \cite{gpt4mini} & 2023.10 \cite{open_ai_table} & OpenAI \\
    GPT-3.5-Turbo & 2023.08 \cite{gpt35turbo_table} & 2021.09 \cite{open_ai_table} & OpenAI \\
    Claude 3.5 Sonnet & 2024.06 \cite{Claude35Sonnet} & 2024.04 \cite{AnthropicModels} & Anthropic \\
    Mixtral 8x7B & 2024.01 \cite{Mistral_table} & -- & Mistral \\
    Llama 3.1 70B & 2024.07 \cite{lama31_table} & 2023.12 \cite{lama31_table2} & Meta \\
    \hline
  \end{tabular}
  \caption{\label{tab:model_cutoffs}
    Release and training data cutoff dates of evaluated models.
  }
\end{table*}

Claude Sonnet Haiku $3$ \cite{ClaudeHaiku} is used to paraphrase data points when necessary. Since Claude Sonnet $3.5$ is included in the evaluation of the methods, the older Haiku $3$ model was used for paraphrasing to avoid potential bias, specifically, the risk that a model might be more likely to recognize its own generated text \cite{Duarte2024}.

\label{sec:results}
\begin{table*}[t]
\caption{\label{tab:table1} AUC-ROC performance of black box methods on the arXiv and Wikipedia datasets using various large language models. Bolded values indicate the highest scores. }
\centering
\small
\begin{tabular}{lcccccccc}
\toprule
Method & \multicolumn{2}{c}{DE-COP} & \multicolumn{2}{c}{Familiarity Ranking} & \multicolumn{2}{c}{Name Cloze Queries} & \multicolumn{2}{c}{Probing} \\
\cmidrule(r){2-3} \cmidrule(r){4-5} \cmidrule(r){6-7} \cmidrule(r){8-9}
Dataset   & arXiv & Wikipedia & arXiv & Wikipedia & arXiv & Wikipedia & arXiv & Wikipedia \\
\midrule
GPT-4o & 0.529 & 0.498 & \textbf{0.568} & 0.528 & \textbf{0.556} & 0.500 &  0.500 & 0.498 \\
GPT4o-mini & 0.419 & 0.502 & 0.491 & 0.519 & 0.496 & 0.519 &  0.500 & 0.500 \\
GPT 3.5 Turbo & 0.482 & 0.460 & 0.487 & 0.509 & 0.548 & 0.500 &  0.500 & 0.502 \\
Claude Sonnet 3.5 & \textbf{0.548} & 0.502 & 0.563 & 0.521 & 0.519 & \textbf{0.538} &  0.500 & 0.502 \\
Mixtral 8x7b & 0.523 & \textbf{0.516} & 0.472 & 0.493 & 0.512 & 0.481 &  0.500 & 0.498 \\
Llama 3.1 70b & - & - & 0.496 & \textbf{0.531} & 0.547 & 0.521 &  \textbf{0.505} & \textbf{0.516} \\
\bottomrule
\label{tab:AUC_Results}
\end{tabular}
\end{table*}

\subsection{Implementation of Novel Method}
The method builds on ranking approaches such as the Likert Scale \cite{Likert}. This approach aims to give the LLM more freedom in expressing how strongly it believes a chunk of text was a piece of the LLM's training corpus. Previous methods often provide the LLM with little to no way of expressing its confidence in a given choice. With a familiarity scale, it will be possible for the LLM to express its confidence while still having restraint in its potential answers. 

A large language model is prompted to rank a set of text chunks based on a given prompt, a rating scale, and the title of the document/article. An example prompt using a simple 1 to 3 ranking scale is shown in Figure~\ref{fig:FR}. The input chunks fall into three categories: the original chunk, a paraphrased version, and randomly selected chunks from the dataset. The original chunk is assigned a label of 1, indicating it is the most likely to appear in the paper; the paraphrased chunks receive a label of 2, and the random chunks are labeled with 0. This part falls under the data manipulation part of the black box methods. The title of the document is either the title of the arXiv paper or the title of the Wikipedia article, depending on the dataset. As described earlier, this acts as a 'hook' to narrow the scope of the LLM answers. 

The LLMs’ task is to rank or score these chunks according to their likelihood of appearing in the target paper. The underlying assumption of this method is that a correct ranking suggests the LLM was exposed to the data during training. To prevent positional bias, the chunk labels are randomly shuffled for each set of chunks. The score is computed based on a perfect match with the correct ordering. If the LLM can output an ordering exactly as in the gold label, the chunk will be marked as a member; otherwise, it will be marked as a non-member. For example, suppose the true label/ordering was [2, 1, 3], representing the paraphrased chunk, original chunks, and random chunk, respectively. In that case, the LLM must generate this exact order to be considered a member data point. The method can be represented as a scoring function:

\begin{equation}
\text{FamiliarityRanking} = \mathcal{M}(t, C, s)
\end{equation}

where $\mathcal{M}$ is the large language model, where $\boldsymbol{t}$ is the title, $\boldsymbol{C}$ is the set of chunks, and $\boldsymbol{s}$ is the ranking scale. In the case of a perfect match between the ranking and the labels of the chunks, the chunk is labeled a 1 for a member and a 0 for a non-member.

\subsection{Evaluation}
The most commonly used metrics for evaluating membership inference attacks are AUC-ROC, constructed from the true positive rate and false positive rate, and accuracy. A low False Positive Rate (FPR) is crucial because it ensures a minimal likelihood of wrongly accusing the owner of a training corpus of containing copyrighted material. A high True Positive Rate (TPR), also known as recall, indicates how many of the copyrighted contents have been successfully identified within the LLM’s training corpus. Therefore, after running the methods on the previously mentioned datasets, the results will be presented in terms of AUC-ROC, true positive rate, and false positive rate. Some experiments will be expressed in terms of accuracy, due to the unfairness of simplifying a complex task into a binary classification problem. 

\subsection{Experiments}
Several experiments were conducted in conjunction with the study to address the research question. The main set of experiments will focus on evaluating the proposed methods across the two datasets. The arXiv dataset contains more challenging and unique content, due to the nature of scientific papers and the chunk selection, making it valuable for assessing how well the methods detect membership in large language models when applied to text that is harder to generalize or reproduce. On the other hand, experiments using Wikipedia articles will evaluate performance on more common and accessible text. Since the content from Wikipedia articles is more readily available, and previous versions of these articles are likely to be included in the training corpus of LLMs, we expect the methods to achieve higher performance on the Wikipedia dataset compared to the arXiv dataset. All methods will be evaluated in terms of AUC-ROC, as well as TRP and FPR. 

The second experiment focuses on evaluating the Probing method. This method relies entirely on the LLM's ability to perform verbatim memorization, meaning it compares the exact text output to the original on a word/token level. Translating this approach into a binary classification task is challenging, as setting an appropriate threshold is highly dependent on the specific context in which the method is applied and what one considers sufficient proof of verbatim memorization. In our setup, the Probing method is evaluated using the Longest Common Subsequence (LCS) metric, which is computed at the word or token level. For all experiments involving member versus non-member classification, the threshold is set to 20 tokens, which is slightly more than 50\% of the suffixes' length, which are approximately 37 tokens. In addition to binary classification, we also evaluate the method using accuracy and LCS scores directly, without converting them into a binary outcome.

The third set of experiments will explore the Familiarity Ranking method in more depth. These experiments will involve evaluating a larger number of chunks and implementing a more unrestricted ranking system. Instead of ranking chunks on a scale from 1 to 3, the large language models will be asked to rate them on a scale from 0 to 10, where 10 indicates the highest likelihood that a chunk belongs to the original paper, and zero the lowest. Accuracy will be used as the evaluation metric, rather than AUC-ROC, since reducing the task to a binary classification problem would not accurately reflect the model's outputs. This expanded scale provides the models with more flexibility in expressing their score, offering better insight into the models' familiarity with each chunk. All paraphrased chunks will be assigned the same ranking, and all of the random chunks will be assigned the same ranking. The experiments are divided into two parts: one using a 3-chunk combination, consisting of the original chunk, one paraphrased chunk, and one random chunk, and another using a 5-chunk combination, which includes two paraphrased chunks, two random chunks, and the original. The increase in the number of chunks will help us see how effectively the method adjusts when dealing with more data points. The LLM's prediction will be considered accurate if it assigns the chunks the appropriate relative ranking, with higher rankings for paraphrased chunks and lower rankings for random ones. 

The final set of experiments will focus on the Name Cloze Queries method and its dependence on the number of proper names/entities included within each chunk. In the original setup \cite{Chang2023}, chunks of 40 to 60 tokens in length containing only a single proper name were selected. This experiment aims to highlight the importance of proper names and entities and investigate how their inclusion affects the performance of the method. 

\begin{table}
  \small
  \centering
  \begin{tabular}{lllll}
    \hline
    \hline Model & Max Tokens & Temp & $top_p$ & $top_k$ \\
    \hline
    \hline GPT4o & 200 & 0.0 & 1.0 & - \\
    \hline GPT4o-mini & 200 & 0.0 & 1.0 & - \\
    \hline GPT 3.5 Turbo & 200 & 0.0 & 1.0 & - \\
    \hline Sonnet 3.5 & 200 & 0.0 & 1.0 &  50 \\
    \hline Mixtral 8x7b & 200 & 0.0 & 1.0 &  50 \\
    \hline Llama 3.1 70b & 200 & 0.0 & 1.0 & 50 \\
    \hline Haiku 3 & 600 & 1.0 & 0.9 &- \\
    \hline
  \end{tabular}
  \caption{\label{tab:Hyperparameter}
    LLMs Hyper parameters
  }
\end{table}

The model hyperparameters used across all methods are listed in Table~\ref{tab:Hyperparameter}. For all models to which the methods are applied, the temperature is set to $0.0$ to ensure the generation of the most likely tokens. In contrast, the model used for paraphrasing, Claude Haiku 3, is assigned a higher temperature to encourage diversity in the generation of the paraphrased chunks.

\begin{table*}[t]
\caption{\label{tab:arxiv_TRP} True positive rate (TPR) and false positive rate (FPR) performance of black box methods on the arXiv dataset using various large language models. Bolded values indicate the highest scores.}
\centering
\small
\begin{tabular}{lcccccccc}
\toprule
\multirow{2}{*}{Method} 
& \multicolumn{2}{c}{DE-COP} 
& \multicolumn{2}{c}{Familiarity} 
& \multicolumn{2}{c}{Name Cloze} 
& \multicolumn{2}{c}{Probing} \\
\cmidrule(r){2-3} \cmidrule(r){4-5} \cmidrule(r){6-7} \cmidrule(r){8-9}
& TPR & FPR & TPR & FPR & TPR & FPR & TPR & FPR \\
\midrule
GPT-4o & 0.745 & 0.688 & 0.459 & 0.323 & 0.612 & 0.500 & 0 & 0 \\
GPT4o-mini & 0.245 & 0.406 & 0.367 & \textbf{0.385} & 0.378 & 0.385 & 0 & 0 \\
GPT-3.5 T & 0.224 & 0.260 & 0.204 & 0.229 & 0.398 & 0.302 & 0 & 0 \\
Sonnet 3.5 & \textbf{0.918} & \textbf{0.823} & \textbf{0.469} & 0.344 & \textbf{0.643} & \textbf{0.604} & 0 & 0 \\
Mixtral 8x7b & 0.245 & 0.198 & 0.173 & 0.229 & 0.367 & 0.344 & 0 & 0 \\
LLaMA 3.1 & - & - & 0.418 & 0.427 & 0.480 & 0.385 & 0.01 & 0 \\
\bottomrule
\end{tabular}
\end{table*}

\begin{table*}[t]
\caption{\label{tab:wikipedia_TRP} True positive rate (TPR) and false positive rate (FPR) performance of black box methods on the Wikipedia dataset using various large language models. Bolded values indicate the highest scores.}
\centering
\small
\begin{tabular}{lcccccccc}
\toprule
\multirow{2}{*}{Method} 
& \multicolumn{2}{c}{DE-COP} 
& \multicolumn{2}{c}{Familiarity} 
& \multicolumn{2}{c}{Name Cloze} 
& \multicolumn{2}{c}{Probing} \\
\cmidrule(r){2-3} \cmidrule(r){4-5} \cmidrule(r){6-7} \cmidrule(r){8-9}
& TPR & FPR & TPR & FPR & TPR & FPR & TPR & FPR \\
\midrule
GPT-4o & 0.751 & 0.756 & 0.394 & 0.338 & 0.803 & \textbf{0.803} & 0.014 & 0.019 \\
GPT4o-mini & 0.319 & 0.315 & 0.362 & 0.324 & 0.690 & 0.653 & 0 & 0 \\
GPT-3.5 T & 0.333 & 0.413 & 0.202 & 0.183 & 0.516 & 0.516 & 0.009 & 0.005 \\
Sonnet 3.5 & \textbf{0.864} & \textbf{0.859} & \textbf{0.526} & \textbf{0.484} & \textbf{0.878} & \textbf{0.803} & 0.019 & 0.014 \\
Mixtral 8x7b & 0.282 & 0.249 & 0.192 & 0.207 & 0.549 & 0.587 & 0.005 & 0.009 \\
LLaMA 3.1 & - & - & 0.404 & 0.343 & 0.700 & 0.657 & 0.061 & 0.028 \\
\bottomrule
\end{tabular}
\end{table*}

\section{Results}
In this section, we examine the performance of various methods across the two datasets and different models. We then introduce an alternative scoring approach for the Familiarity Ranking method and analyze its results across varying numbers of chunks. Next, we examine the impact of proper entities on Name Cloze Queries and how their quantity affects the accuracy across different models. Finally, we present alternative metrics for evaluating the Probing method.

\subsection{Performance of the different methods}


The first set of experiments aims to assess the overall performance of the methods. Since the primary objective of the methods is to distinguish between member and non-member data points within the LLM's training corpus, the results are evaluated as a binary classification task using the AUC-ROC metric. The outcomes are summarized in Table~\ref{tab:AUC_Results}.

From the overall results, we observe that all methods struggle to differentiate between member and non-member data points, with AUC-ROC scores being around 0.5. While some methods, such as Familiarity Ranking on the arXiv dataset using GPT-4o and Claude Sonnet 3.5, achieve slightly higher scores of around 0.6, this is still insufficient to demonstrate that the methods can reliably distinguish between member and non-member chunks. The results for DE-COP using Llama 3.1 are not presented, as the model struggled with this task and approximately 20\% of its answers were null values for each run. Most of the null values were different in each run. The results of DE-COP using Llama 3.1 are presented in the appendix.

Table~\ref{tab:arxiv_TRP} and~\ref{tab:wikipedia_TRP} provide a more detailed view of the results by breaking down the AUC-ROC into true positive rate (TPR) and false positive rate (FPR). The similarity between TPR and FPR across all models suggests that they perform the task equally for both member and non-member data points. We also observe that results on the Wikipedia dataset (Table~\ref{tab:wikipedia_TRP}) are generally better, particularly for older models.

Interestingly, newer models such as GPT-4o, Claude Sonnet 3.5, and LLaMA 3.1 tend to show both higher TPR and FPR. This indicates that these models possess stronger reasoning and generalization abilities, allowing them to infer correct answers even for unseen data. Furthermore, some methods underperform on specific models; for example, DE-COP appears to struggle with LLaMA 3.1, implying the importance of testing MIA methods on multiple families of models. The Probing method also yields weak results across all models, due to the difficulty of translating the task into a binary classification problem. Section 4.4 explores alternative metrics for evaluating Probing in more depth.

On average, the arXiv dataset shows lower TPR and FPR values compared to the dataset constructed from Wikipedia articles. This is an expected outcome, as Wikipedia content tends to be more accessible and easier to reason about than the more technical and complex content found in arXiv papers. Additionally, the Wikipedia
articles are updated versions of earlier entries, suggesting that LLMs may be able to infer non-member versions using knowledge acquired from earlier editions. Interestingly, this trend does not hold for our Familiarity Ranking method. For this method, both TPR and FPR are generally higher on the arXiv dataset across all models, except for Claude Sonnet 3.5, which has a higher TPR and FPR for the Wikipedia dataset.

\subsection{Different ranking scale for Familiarity Ranking}

The Familiarity Ranking method introduced in this paper allows for different ranking scales, as discussed earlier. In the current setup, we use a 0 to 10 scale, providing LLMs with more flexibility in expressing familiarity. The results are presented in Table~\ref{tab:FR_acc}.

\begin{table}[t]
\caption{\label{tab:table1} The accuracy of the models over member(M) and non-member(non-M) chunks of data from the arXiv dataset when running Familiarity Ranking with 0 to 10 rating.}
\centering
\small
\begin{tabular}{lcccc}
\toprule
 & \multicolumn{2}{c}{3-chunks} & \multicolumn{2}{c}{5-chunks} \\
\cmidrule(r){2-3} \cmidrule(r){4-5}
 & M & non-M & M & non-M  \\
\midrule
GPT-4o & 0.680 & 0.743 & 0.779 & 0.788  \\
GPT4o-mini & 0.633 & 0.642 & 0.711 & 0.747  \\
GPT-3.5 Turbo & 0.571 & 0.573 & 0.697 & 0.677  \\
Sonnet 3.5 & 0.742 & 0.760 & 0.850 & 0.816  \\
Llama 3.1 & 0.537 & 0.493 & 0.728 & 0.740  \\
\bottomrule
\label{tab:FR_acc}
\end{tabular}
\end{table}

The results show that the accuracies for member and non-member data points are very close, highlighting the reasoning capabilities of the LLMs. Across all models, the original and paraphrased chunks consistently received high scores (typically between 8 and 10), while random chunks were assigned much lower scores (usually between 0 and 3). This demonstrates that the LLMs can reason effectively, even with only the title as context. This pattern is further supported by the improved accuracy observed in the 5-chunk setup compared to the 3-chunk setup. The likely explanation is that the LLMs are particularly effective at identifying and excluding the random chunks, which increases overall accuracy as more distraction chunks are added. 

\subsection{Impact of proper entities on Name Cloze Queries}

In the original Name Cloze Queries methodology \cite{Chang2023}, chunks of 40 to 60 tokens containing only a single proper name were used. To investigate the impact of the number of proper names on the accuracy of Name Cloze Queries, we conducted an experiment in which all chunks were masked regardless of the number of proper names. We used the same arXiv data as for previous experiments. To avoid simplifying the task as a binary classification problem, the results are instead presented using accuracy as the evaluation metric. Each correctly inferred token from a chunk is counted and divided by the total number of chunks, and the resulting accuracies are then averaged across all chunks. The results of this experiment are presented in Table~\ref{tab:NCQ}.


\begin{table}[h]
  \centering
  \begin{tabular}{llll}
    \hline Model & member & non-member  \\
    \hline GPT4o & 0.395 & 0.316  \\
    GPT4o-mini & 0.194 & 0.156 \\
    GPT 3.5 Turbo & 0.242 & 0.223 \\
    Claude Sonnet 3.5 & 0.458 & 0.362 \\
    Mixtral 8x7b & 0.294 & 0.215   \\
    Llama 3.1 70b & 0.310 & 0.238  \\
    \hline
  \end{tabular}
  \caption{\label{tab:NCQ}
    The accuracy of the models over member and non-member chunks of data for the arXiv dataset when running the Name Cloze Queries with all of the tokens in the chunk masked.
  }
\end{table}

As shown in Table~\ref{tab:NCQ}, the accuracy for member data points is consistently higher across all models. The difference is minor for older models, while newer models show a larger gap. These findings provide the most substantial evidence of memorization capabilities among all the evaluated experiments, as member accuracies are consistently higher; for newer models, the difference is especially significant, reaching approximately 10\%.

\begin{figure*}
\centering
\includegraphics[width=11cm, height=6cm]{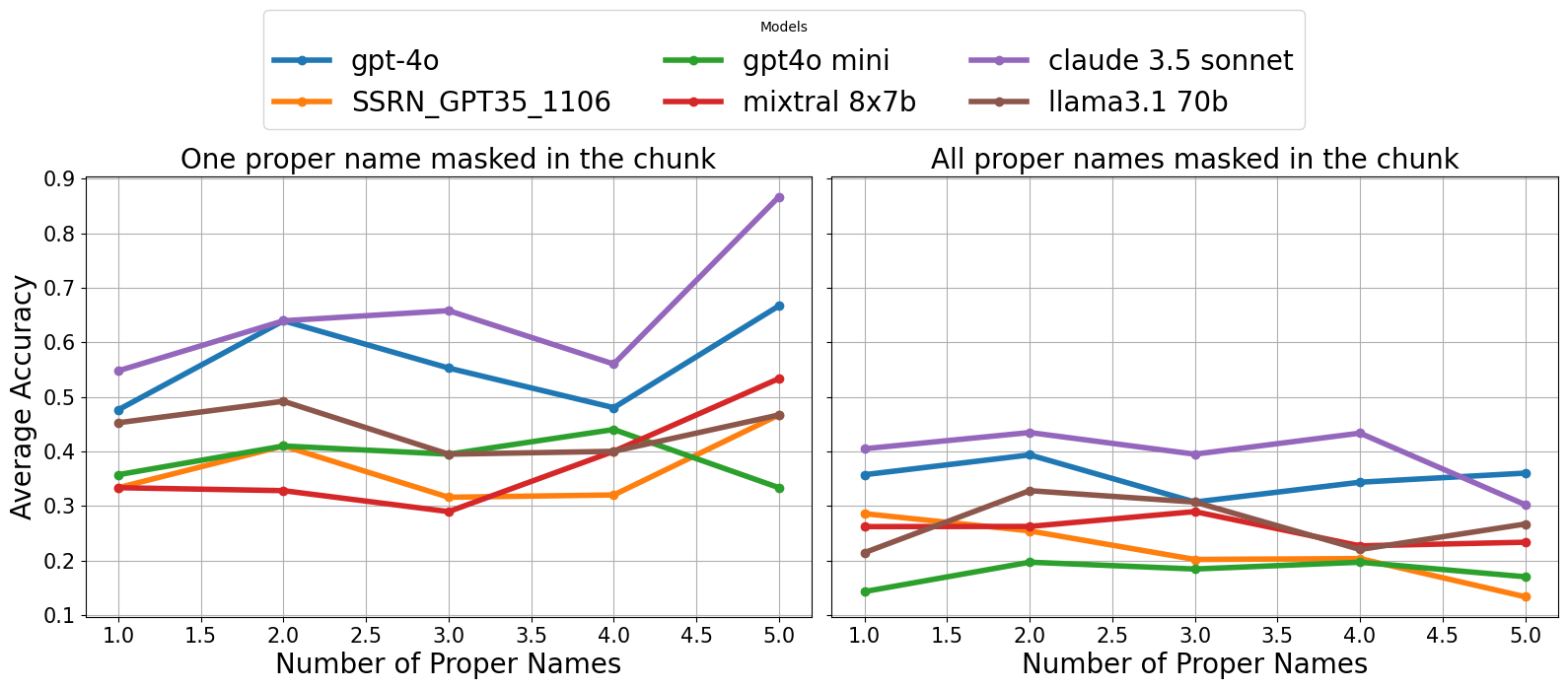}
\caption{Average accuracy of Name Cloze Queries per different number of name entities present in the chunk.}
\label{fig:NCQ}
\end{figure*}

To further evaluate the difference between masking all tokens versus masking only one, we present a side-by-side comparison of the average accuracies of the Name Cloze Queries across chunks with different numbers of proper names. While the arXiv dataset contains a few chunks with more than five proper names, they do not provide a sufficiently large sample size to be included in this evaluation. The results are shown in Figure~\ref{fig:NCQ}.

From the graphs, we can observe that the overall accuracy is consistently higher when only one proper name is masked, compared to when all proper names in the chunk are masked, regardless of the number of proper names in the chunk. Interestingly, the best-performing model, Claude Sonnet 3.5, still outperforms the two weakest models, Mixtral 8x7B and GPT-3.5 Turbo, even when only one proper name is masked. This behavior is somewhat unexpected, as the accuracy for chunks with only one proper name should be the same across setups. A likely explanation of this is the slight change in the prompt, which tells the LLM about the possibility of multiple proper names being present, potentially influencing its reasoning. Giving the same prompt (the same as the one used in single-token masking), however, also hinders the results, yielding similar accuracies. 

Furthermore, we observe that when one proper name is masked, the accuracy improves for all models except GPT4o-mini, particularly when there are five proper names in the chunk. On the other hand, when all proper names are masked, the accuracy remains relatively stable across all models, regardless of the number of proper names in the chunk.

\subsection{Longest Common Sequence Probing}

From the results in Table~\ref{tab:arxiv_TRP} and~\ref{tab:wikipedia_TRP}, we can see that the Probing method does not achieve strong results. However, due to the nature of the method, which focuses strictly on verbatim memorization, it is difficult to compare it to other methods. In calculating the AUC-ROC, the threshold for classifying a data point as a member was set at 20 tokens, which corresponds to more than 50\% of the suffix, as the suffixes contain approximately 37 tokens. Table~\ref{tab:Wiki_LCS} and Table~\ref{tab:arxiv_LCS} compare the AUC-ROC score with ACC and Longest Common Sequence (LCS), which are different ways of evaluating the Probing method. 


\begin{table}[t]
\caption{\label{tab:table1}Model performance over member (M) and non-member (non-M) chunks of arXiv with different metrics to evaluate Probing. Longest Common Sequence (LCS) is rounded to the nearest full token.}
\label{tab:arxiv_LCS}
\centering
\small
\begin{tabular}{lcccccc}
\toprule
Model & AUC & \multicolumn{2}{c}{ACC} & \multicolumn{2}{c}{LCS} \\
      &     & M & Non-M & M & Non-M \\
\midrule
GPT-4o        & 0.500 & 0.339 & 0.286 & 8 & 6 \\
GPT4o-mini    & 0.500 & 0.301 & 0.267 & 7 & 6 \\
GPT-3.5 T     & 0.500 & 0.237 & 0.201 & 5 & 4 \\
Sonnet 3.5    & 0.500 & 0.256 & 0.216 & 5 & 4 \\
Mixtral 8x7b  & 0.500 & 0.377 & 0.331 & 9 & 8 \\
Llama 3.1     & 0.505 & 0.293 & 0.254 & 7 & 6 \\
\bottomrule
\end{tabular}
\end{table}


\begin{table}[t]
\caption{Model performance over member (M) and non-member (non-M) chunks of Wikipedia with different metrics to evaluate Probing. Longest Common Sequence (LCS) is rounded to the nearest full number.}
\label{tab:Wiki_LCS}
\centering
\small
\begin{tabular}{lcccccc}
\toprule
Model & AUC & \multicolumn{2}{c}{ACC} & \multicolumn{2}{c}{LCS} \\
      &     & M & Non-M & M & Non-M \\
\midrule
GPT-4o        & 0.498 & 0.360 & 0.366 & 8 & 9 \\
GPT4o-mini    & 0.500 & 0.313 & 0.314 & 7 & 7 \\
GPT-3.5 T     & 0.502 & 0.287 & 0.273 & 6 & 6 \\
Sonnet 3.5    & 0.502 & 0.335 & 0.325 & 5 & 4 \\
Mixtral 8x7b  & 0.498 & 0.394 & 0.389 & 9 & 10 \\
Llama 3.1     & 0.516 & 0.308 & 0.304 & 8 & 7 \\
\bottomrule
\end{tabular}
\end{table}

The accuracy results in both tables align with findings from previous experiments, which indicate that there is no significant difference between member and non-member data points. This suggests that the Probing method also struggles to distinguish between the two, as any slight advantage in accuracy may come from factors unrelated to memorization. The factors might include hallucination or generalization. This trend holds across both datasets, Wikipedia and arXiv.

When examining the Longest Common Subsequence (LCS) calculated at the token level, we observe that member chunks in the arXiv dataset tend to have slightly longer common sequences on average, typically one additional token, compared to the non-member chunks. However, in the Wikipedia dataset, non-member chunks show equal or even higher LCS values, as seen with the Mixtral 8x7B model. This could be due to the nature of the dataset, where multiple versions of similar articles from different years may lead to shared phrasing even among non-member chunks.

\section{Discussion}
\label{sec:discussion}

\subsection{Analysis of Findings}
State-of-the-art methods for membership inference attacks (MIA) under the black-box assumption have evaluated their performance on datasets containing passages from popular books \cite{Duarte2024, Karamolegkou2023, Chang2023}. Books, similar to Wikipedia articles and arXiv papers, are a popular source for training materials for large language models (LLMs). However, books often serve as widely recognized materials, frequently used and copied across multiple sites over many years.

Our study includes more challenging datasets, and the results show that models struggle to distinguish between member and non-member chunks. This difficulty is likely tied to findings in prior work, which indicate that increased repetition of a text in the training data significantly raises the likelihood of an LLM reproducing it \cite{Carlini2023}. This is especially evident in the arXiv dataset, where the chunks are extracted from the most likely unique sections of the papers. This is illustrated by the similar true positive rate (TPR) and false positive rate (FPR) observed across the models. In practice, a significantly higher TPR than FPR would indicate memorization. Still, the similar TPR and FPR values suggest that the models, especially the newer ones, are capable of logical reasoning and correctly answer for both member and non-member data points. For example, as shown in Table~\ref{tab:AUC_Results}, Claude Sonnet 3.5 achieves a TPR of over 0.9 and an FPR of over 0.82 under the DE-COP method on the arXiv dataset. This indicates that the model is answering correctly, not due to memorization, but likely through generalization and reasoning. These results highlight the increasing difficulty of detecting membership as LLMs become more capable of reasoning over unseen data. The higher TPR values observed in some methods are insufficient to indicate true memorization, especially given the inconsistency of these differences across models. In some cases, the FPR is even higher than the TPR, further weakening the argument for reliable black-box memorization detection methods. 

The Familiarity Ranking method developed in this paper does not outperform the other state-of-the-art methods presented. While it achieves the highest AUC-ROC score among all methods and models, under the GPT-4o model, this result is not strong enough to conclude that the method is effective at distinguishing member from non-member data points, even for GPT-4o. Additionally, the TPR and FPR values for Familiarity Ranking are consistently the lowest among the tested methods (excluding the Probing method), suggesting that models struggle to differentiate between the original and paraphrased chunks. This is somewhat surprising, especially when compared to the DE-COP method, where models, particularly the newer ones, have high TPR and FPR, even when tasked with distinguishing between an original chunk and three paraphrased versions. This suggests that the addition of one random chunk into the choice pool confuses the LLM, making the task more challenging. 

It is also essential to highlight the variation in performance across different models. Newer models consistently perform better across all methods, highlighting the importance of evaluating membership inference techniques on the most recent LLMs. As language models continue to improve in generalization and reasoning capabilities, it becomes increasingly challenging to evaluate these black-box methods and to demonstrate that LLMs are reproducing text through memorization reliably. It was also found during the experiments that the prompts used in the original method's studies have become less effective with the newer models. For example, the newer OpenAI models, GPT-4o and GPT4o-mini, refuse to produce results giving the original Probing prompt asking for verbatim text; they instead propose to provide a summary, safeguarding the potential claims of verbatim memorization, unlike the older GPT 3.5 Turbo, which tries to give you the verbatim text. A jailbreaking prompt \cite{jailbreak} had to be used to obtain the results. This highlights another aspect of the growing difficulty in using black-box methods for membership inference attacks on LLMs, in addition to the increasing reasoning and generalization capabilities of these models. 

\subsection{Analysis of Members and Non-members}

\begin{figure}[h]
\includegraphics[width=7cm, height=6cm]{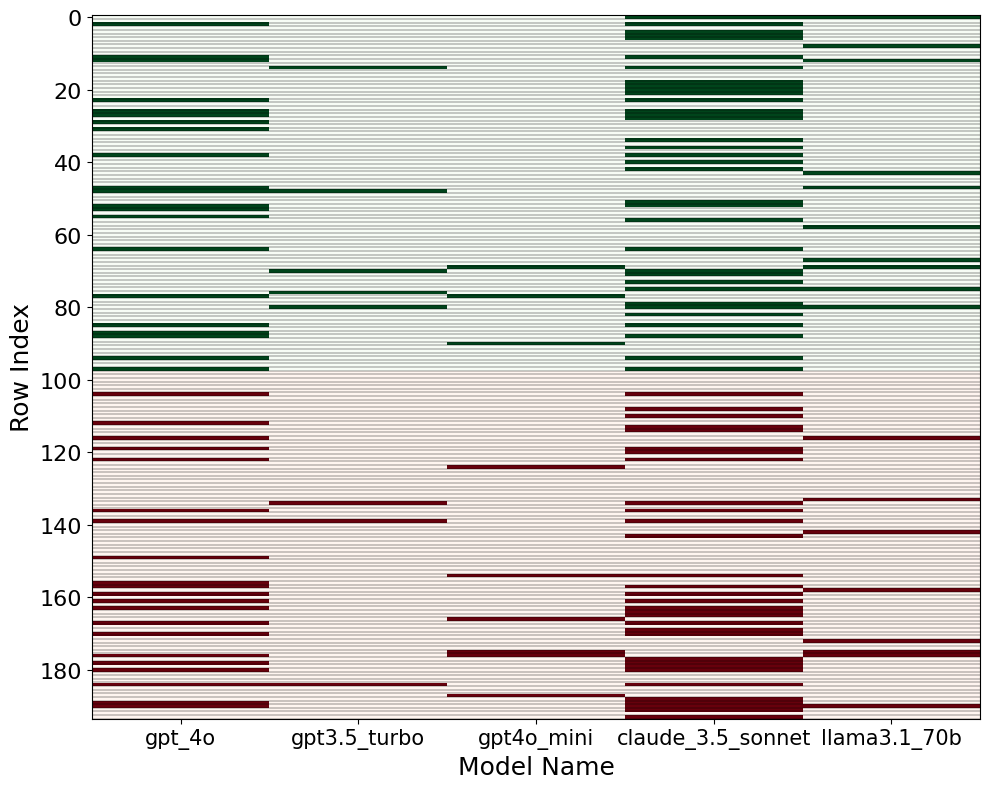}
\caption{Heatmap representing articles in the arXiv dataset where all of the methods suggest memorization. The green colors represent potential member articles, and the red colors represent non-member articles.}
\label{fig:heatmap}
\end{figure}

The heatmap visible in Figure~\ref{fig:heatmap}, highlights the dataset indices where all three methods, DE-COP, FR, and NCQ, agree on a document being a member. The NCQ, in this case, is the original version with only one proper name masked, and FR uses a 0-10 scale. As we can see, the same trend follows as in Table~\ref{tab:arxiv_TRP} and~\ref{tab:wikipedia_TRP}, where the models GPT-4o and Claude Sonnet 3.5 present the highest number of recognized chunks. It is also evident from the heatmap that all the models have a similar number of member and non-member chunks recognized, supporting our previous results. It is also interesting to note that no articles are consistently recognized as members by more than two LLMs, even those from the same family of LLMs. It further shows the importance of testing black-box MIAs on multiple LLMs. The recognized chunks themselves also do not display any pattern in their structure or semantics. However, examining the results with our method, it can be seen that older models exhibit more confidence in selecting the correct chunk than the newer LLMs. GPT 3.5 Turbo shows the most confidence, even when wrong, being the oldest model. Suggesting that more advanced LLMs may rely more on reasoning when faced with tasks and show more uncertainty, making it harder to prove memorization in newer LLMs, while older models are more likely to rely on memorization.  

The assumption about non-member data chunks can be considered a strong assumption, as it is implausible that a piece of data has been used in the training of an LLM past its cut-off date. The assumption of the member data chunk, however, is very loose, as not only might the document from which the chunk is derived not have been in the training corpus, but the chunk itself may also not have been a part of the training corpus. That being said, the articles marked as members by all of the methods should not be disregarded. They may help create a future dataset, as member chunks, where a complete collection of them can serve as a benchmark for new methods, since they are regarded as members by all current methods. All of the articles from which chunks have been marked as members by all three methods can be seen in the appendix. 

\subsection{Limitations}
The limitations of this study, first of all, consist of the assumptions made about the methods and datasets. More precisely, the assumption about the selected data points being members or non-members and the assumption about proving memorization given the output of the methods. The assumption about the data points is necessary, as unfortunately, the training data of the LLMs is not publicly available, and golden labels must be inferred based on the assumption that the LLMs contain or do not contain a given piece of data. The other assumption, regarding the methods, is necessary for the evaluation. However, one can disagree with one of the methods' methodologies and suggest that it does not prove memorization. 

Another limitation of this study is the scope of evaluated methods and models. While the results indicate that performance varies substantially across different LLMs, only a subset of available models and black-box methods were assessed. Expanding the evaluation to additional models, methods, and data sources would provide a more comprehensive understanding of the robustness and generalizability of the findings. Furthermore, increasing the number of data points could improve statistical reliability.

\subsection{Future Work}
One possible direction for further study is a supplementary evaluation of the novel method. The method we developed can be tested with a broader range of scoring metrics, as LLMs may infer different results and, with different scales, provide more depth to their answers. The results can also be analyzed in more depth, examining each of the chunks and understanding how and why the LLMs assigned a specific score to each one. This study, along with all the previous studies discussed in this paper, focuses on memorizing chunks; an analysis of memorizing whole documents might be a direction for future research. 

\section{Conclusion}
\label{sec:conclusion}

This study demonstrates that black-box membership inference attacks may not be a viable way of inferring membership from large language models. The results show that modern LLMs demonstrate high levels of reasoning and can infer results from data points they have never seen before, achieving results that are barely worse than those from data points they have potentially seen in their training data. These results are, however, under several assumptions made during this study and might need to be further evaluated with access to the training data of an advanced LLM. During the study, it was also demonstrated that some methods, particularly the Probing method, struggled to obtain results with the original prompt, highlighting possible safeguards implemented in newer LLMs to prevent the reproduction of copyrighted material. 

The novel method introduced in this paper faces the same challenges as the other methods, namely, the difficulty of distinguishing true memorization from reasoning. Instead of identifying memorized data points, the method highlights the advanced reasoning abilities of newer LLMs and the difficulty of obtaining clear results using black-box MIA. This is evident in the models' consistent assignment of low scores to random chunks, while giving high scores to both the original and paraphrased chunks, suggesting that the models rely more on reasoning and semantics rather than pure memorization.

Even though this study shows that black-box methods struggle to infer membership in large language models, other types of membership inference attacks (gray box and white box) should not be disregarded. While these methods require more access to the LLMs, they have shown promising results, as stated in the related works section. However, a key limitation is that LLMs allowing for gray box or white box access are often not the most recent ones or community-built, which restricts the ability to evaluate these methods on state-of-the-art LLMs. As this study has shown, newer LLMs tend to exhibit better reasoning capabilities. Since this field of study is relatively new, the older gray and white box methods may not work as well with the newer LLMs. 

\onecolumn

\bibliography{latex/references}

@misc{Chang2023,
    title = {{Speak, Memory: An Archaeology of Books Known to ChatGPT/GPT-4}},
    author = {Kent K. Chang and Mackenzie Cramer and Sandeep Soni and David Bamman},
    year = {2023},
    archivePrefix = {arXiv},
    url = {https://arxiv.org/abs/2305.00118}
}

@misc{Karamolegkou2023,
    title = {{Copyright Violations and Large Language Models}},
    author = {Antonia Karamolegkou and Jiaang Li and Li Zhou and Anders Søgaard},
    year = {2023},
    archivePrefix = {arXiv},
    url = {https://arxiv.org/abs/2310.13771}
}

@misc{Hartmann2023,
    title = {{SoK: Memorization in General-Purpose Large Language Models}},
    author = {Valentin Hartmann and Anshuman Suri and Vincent Bindschaedler and David Evans and Shruti Tople and Robert West},
    year = {2023},
    archivePrefix = {arXiv},
    url = {https://arxiv.org/abs/2310.18362}
}

@misc{Deng2023,
    title = {{Investigating Data Contamination in Modern Benchmarks for Large Language Models}},
    author = {Chunyuan Deng and Yilun Zhao and Xiangru Tang and Mark Gerstein and Arman Cohan},
    year = {2023},
    archivePrefix = {arXiv},
    url = {https://arxiv.org/abs/2311.09783}
}

@misc{Duan2024,
    title = {{Do Membership Inference Attacks Work on Large Language Models?}},
    author = {Michael Duan and Anshuman Suri and Niloofar Mireshghallah and Sewon Min and Weijia Shi and Luke Zettlemoyer and Yulia Tsvetkov and Yejin Choi and David Evans and Hannaneh Hajishirzi},
    year = {2024},
    archivePrefix = {arXiv},
    url = {https://arxiv.org/abs/2402.07841}
}

@misc{Duarte2024,
    title = {{DE-COP: Detecting Copyrighted Content in Language Models Training Data}},
    author = {André V. Duarte and Xuandong Zhao and Arlindo L. Oliveira and Lei Li},
    year = {2024},
    archivePrefix = {arXiv},
    url = {https://arxiv.org/abs/2402.09910}
}

@misc{Dong2024,
    title = {{Generalization or Memorization: Data Contamination and Trustworthy Evaluation for Large Language Models}},
    author = {Yihong Dong and Xue Jiang and Huanyu Liu and Zhi Jin and Bin Gu and Mengfei Yang and Ge Li},
    year = {2024},
    archivePrefix = {arXiv},
    url = {https://arxiv.org/abs/2402.15938}
}

@misc{Shi2024,
    title = {{Detecting Pretraining Data from Large Language Models}},
    author = {Weijia Shi and Anirudh Ajith and Mengzhou Xia and Yangsibo Huang and Daogao Liu and Terra Blevins and Danqi Chen and Luke Zettlemoyer},
    year = {2024},
    archivePrefix = {arXiv},
    url = {https://arxiv.org/abs/2310.16789}
}

@misc{Stoehr2024,
    title = {{Localizing Paragraph Memorization in Language Models}},
    author = {Niklas Stoehr and Mitchell Gordon and Chiyuan Zhang and Owen Lewis},
    year = {2024},
    archivePrefix = {arXiv},
    url = {https://arxiv.org/abs/2403.19851}
}

@misc{Nakka2024,
    title = {{PII-Compass: Guiding LLM training data extraction prompts towards the target PII via grounding}},
    author = {Krishna Kanth Nakka and Ahmed Frikha and Ricardo Mendes and Xue Jiang and Xuebing Zhou},
    year = {2024},
    archivePrefix = {arXiv},
    url = {https://arxiv.org/abs/2407.02943}
}

@misc{Zhang2024[SURP],
    title = {{Adaptive Pre-training Data Detection for Large Language Models via Surprising Tokens}},
    author = {Anqi Zhang and Chaofeng Wu},
    year = {2024},
    archivePrefix = {arXiv},
    url = {https://arxiv.org/abs/2407.21248}
}

@misc{Wang2024[CON-Recall],
    title = {{Con-ReCall: Detecting Pre-training Data in LLMs via Contrastive Decoding}},
    author = {Cheng Wang and Yiwei Wang and Bryan Hooi and Yujun Cai and Nanyun Peng and Kai-Wei Chang},
    year = {2024},
    archivePrefix = {arXiv},
    url = {https://arxiv.org/abs/2409.03363}
}

@misc{Zhang2024[FPR_def],
    title = {{Membership Inference Attacks Cannot Prove that a Model Was Trained On Your Data}},
    author = {Jie Zhang and Debeshee Das and Gautam Kamath and Florian Tramèr},
    year = {2024},
    archivePrefix = {arXiv},
    url = {https://arxiv.org/abs/2409.19798}
}

@misc{Satvaty2024,
    title = {{Undesirable Memorization in Large Language Models: A Survey}},
    author = {Ali Satvaty and Suzan Verberne and Fatih Turkmen},
    year = {2024},
    archivePrefix = {arXiv},
    url = {https://arxiv.org/abs/2410.02650}
}

@misc{Kim2024,
    title = {{Detecting Training Data of Large Language Models via Expectation Maximization}},
    author = {Gyuwan Kim and Yang Li and Evangelia Spiliopoulou and Jie Ma and Miguel Ballesteros and William Yang Wang},
    year = {2024},
    archivePrefix = {arXiv},
    url = {https://arxiv.org/abs/2410.07582}
}

@misc{Fu2024,
    title = {{Does Data Contamination Detection Work (Well) for LLMs? A Survey and Evaluation on Detection Assumptions}},
    author = {Yujuan Fu and Ozlem Uzuner and Meliha Yetisgen and Fei Xia},
    year = {2024},
    archivePrefix = {arXiv},
    url = {https://arxiv.org/abs/2410.18966}
}

@misc{Oren2024,
    title = {{Proving Test Set Contamination in Black Box Language Models}},
    author = {Yonatan Oren and Nicole Meister and Niladri Chatterji and Faisal Ladhak and Tatsunori B. Hashimoto},
    year = {2024},
    archivePrefix = {arXiv},
    url = {https://arxiv.org/abs/2310.17623}
}

@misc{Xie2024[LiMem],
    title = {{On Memorization of Large Language Models in Logical Reasoning}},
    author = {Chulin Xie and Yangsibo Huang and Chiyuan Zhang and Da Yu and Xinyun Chen and Bill Yuchen Lin and Bo Li and Badih Ghazi and Ravi Kumar},
    year = {2024},
    archivePrefix = {arXiv},
    url = {https://arxiv.org/abs/2410.23123}
}

@misc{Song2024,
    title = {{Both Text and Images Leaked! A Systematic Analysis of Multimodal LLM Data Contamination}},
    author = {Dingjie Song and Sicheng Lai and Shunian Chen and Lichao Sun and Benyou Wang},
    year = {2024},
    archivePrefix = {arXiv},
    url = {https://arxiv.org/abs/2411.03823}
}

@misc{EducatingSilicon2024,
    author = {{Educating Silicon}},
    title = {{How Much LLM Training Data is There in the Limit?}},
    year = {2024},
    url = {https://www.educatingsilicon.com/2024/05/09/how-much-llm-training-data-is-there-in-the-limit/},
    note = {Accessed: 2025-02-22}
}

@misc{NYT2023,
    author = {{Michael M. Grynbaum and Ryan Mac}},
    title = {{New York Times Sues OpenAI and Microsoft Over Copyright}},
    year = {2023},
    url = {https://www.nytimes.com/2023/12/27/business/media/new-york-times-open-ai-microsoft-lawsuit.html},
}

@misc{Zhang2024_Min-kpp,
    title = {{Min-K++: Improved Baseline for Detecting Pre-Training Data from Large Language Models}},
    author = {Jingyang Zhang and Jingwei Sun and Eric Yeats and Yang Ouyang and Martin Kuo and Jianyi Zhang and Hao Frank Yang and Hai Li},
    year = {2024},
    archivePrefix = {arXiv},
    url = {https://arxiv.org/abs/2404.02936}
}

@misc{Xie2024[RECALL],
    title = {{ReCaLL: Membership Inference via Relative Conditional Log-Likelihoods}},
    author = {Roy Xie and Junlin Wang and Ruomin Huang and Minxing Zhang and Rong Ge and Jian Pei and Neil Zhenqiang Gong and Bhuwan Dhingra},
    year = {2024},
    archivePrefix = {arXiv},
    url = {https://arxiv.org/abs/2406.15968}
}

@misc{OpenAIModels,
    author = {{OpenAI}},
    title = {{Models}},
    year = {2025},
    url = {https://platform.openai.com/docs/models},
}

@misc{AnthropicModels,
    author = {{Anthropic}},
    title = {{All models overview}},
    year = {2025},
    url = {https://docs.anthropic.com/en/docs/about-claude/models/all-models},
}

@misc{MetaModels,
    author = {{Meta}},
    title = {{Llama 3 models}},
    year = {2025},
    url = {https://www.llama.com/models/llama-3/},
}

@misc{LatestEval,
    title = {{LatestEval: Addressing Data Contamination in Language Model Evaluation through Dynamic and Time-Sensitive Test Construction}},
    author = {Yucheng Li and Frank Guerin and Chenghua Lin},
    year = {2023},
    archivePrefix = {arXiv},
    url = {https://arxiv.org/abs/2312.12343}
}

@misc{mistralmodels,
    author = {{Mistral AI}},
    title = {{Mistral AI}},
    year = {2025},
    url = {https://mistral.ai/},
}

@misc{Gpt4,
    title = {{GPT-4o System Card}},
    author = {OpenAI},
    year = {2024},
    archivePrefix = {arXiv},
    url = {https://arxiv.org/pdf/2410.21276}
}

@misc{Mixtral,
    title = {{Mixtral of Experts}},
    author = {Albert Q. Jiang and Alexandre Sablayrolles and Antoine Roux and Arthur Mensch and Blanche Savary and Chris Bamford and Devendra Singh Chaplot and Diego de las Casas and Emma Bou Hanna and Florian Bressand and Gianna Lengyel and Guillaume Bour and Guillaume Lample and Lélio Renard Lavaud and Lucile Saulnier and Marie-Anne Lachaux and Pierre Stock and Sandeep Subramanian and Sophia Yang and Szymon Antoniak and Teven Le Scao and Théophile Gervet and Thibaut Lavril and Thomas Wang and Timothée Lacroix and William El Sayed},
    year = {2024},
    archivePrefix = {arXiv},
    url = {https://arxiv.org/pdf/2401.04088}
}

@misc{Likert,
    title = {{Likert Scale: Explored and Explained}},
    author = {Ankur Joshi and Saket Kale and Satish Chandel and Dinesh Kumar Pal},
    year = {2015},
    url = {https://www.researchgate.net/publication/276394797_Likert_Scale_Explored_and_Explained}
}

@misc{Claude35Sonnet,
    author = {{Anthropic}},
    title = {{Claude 3.5 Sonnet}},
    year = {2025},
    url = {https://www.anthropic.com/news/claude-3-5-sonnet},
}

@misc{gpt4mini,
    author = {{OpenAI}},
    title = {{GPT-4o mini: advancing cost-efficient intelligence}},
    yea = {2024},
    url = {https://openai.com/index/gpt-4o-mini-advancing-cost-efficient-intelligence/},
}

@misc{Hallucination,
    title = {{A Survey on Hallucination in Large Language Models: Principles, Taxonomy, Challenges, and Open Questions}},
    author = {Lei Huang and Weijiang Yu and Weitao Ma and Weihong Zhong and Zhangyin Feng and Haotian Wang and Qianglong Chen and Weihua Peng and Xiaocheng Feng and Bing Qin and Ting Liu},
    year = {2023},
    archivePrefix = {arXiv},
    url = {https://arxiv.org/abs/2311.05232}
}

@misc{Artificial_Analysis,
    author = {{Artificial Analysis}},
    title = {{OpenAI: Models Intelligence, Performance \& Price}},
    url = {https://artificialanalysis.ai/providers/openai},
}

@misc{Mistral_table,
    author = {{Mistral}},
    title = {{Mixtral of experts}},
    year = {2023},
    url = {https://mistral.ai/news/mixtral-of-experts},
}

@misc{lama31_table,
    author = {{Meta}},
    title = {{Introducing Llama 3.1: Our most capable models to date}},
    year = {2024},
    url = {https://ai.meta.com/blog/meta-llama-3-1/},
}

@misc{lama31_table2,
    author = {{Meta}},
    title = {{Llama-3.1-70B}},
    url = {https://huggingface.co/meta-llama/Llama-3.1-70B},
}

@misc{open_ai_table,
    author = {{Azure OpenAI}},
    title = {{Azure OpenAI in Azure AI Foundry Models}},
    year = {2025},
    url = {https://learn.microsoft.com/en-us/azure/ai-services/openai/concepts/models?tabs=global-standard%2Cstandard-chat-completions#gpt-35-turbo-model-availability},
}

@misc{gpt4o_table,
    author = {{OpenAI}},
    title = {{Hello GPT-4o}},
    year = {2024},
    url = {https://openai.com/index/hello-gpt-4o/},
}

@misc{gpt35turbo_table,
    author = {{OpenAI}},
    title = {{GPT-3.5 Turbo fine-tuning and API updates}},
    year = {2023},
    url = {https://openai.com/index/gpt-3-5-turbo-fine-tuning-and-api-updates/},
}

@misc{ClaudeHaiku,
    author = {{Anthropic}},
    title = {{Introducing the next generation of Claude}},
    year = {2024},
    url = {https://www.anthropic.com/news/claude-3-family},
}

@misc{Carlini2023,
    title = {{Extracting Training Data from Diffusion Models}},
    author = {Nicholas Carlini and Jamie Hayes and Milad Nasr and Matthew Jagielski and Vikash Sehwag and Florian Tramèr and Borja Balle and Daphne Ippolito and Eric Wallace},
    year = {2023},
    archivePrefix = {arXiv},
    url = {https://arxiv.org/abs/2301.13188}
}

@misc{jailbreak,
    title = {{Don’t Listen To Me: Understanding and Exploring Jailbreak Prompts of Large Language Models}},
    author = {Zhiyuan Yu and Xiaogeng Liu and Shunning Liang and Zach Cameron and Chaowei Xiao and Ning Zhang},
    year = {2024},
    archivePrefix = {arXiv},
    url = {https://arxiv.org/html/2403.17336v1}
}

@misc{LLMsurvey,
    title = {{Large Language Models: A Survey}},
    author = {Shervin Minaee and Tomas Mikolov and Narjes Nikzad and Meysam Chenaghlu and Richard Socher and Xavier Amatriain and Jianfeng Gao},
    year = {2024},
    archivePrefix = {arXiv},
    url = {https://arxiv.org/abs/2402.06196}
}

\appendix

\section{Prompts}
\label{sec:apx:first_appendix}

\subsection{Probing}

\begin{figure}[H]
\noindent\fbox{%
  \parbox{\textwidth}{%
    \hspace*{0em}In order to test the capabiliites of Large Language Models, you are asked to complete the following task. You are given a title and a prefix of a popular arXiv paper. Your task is to complete the passage that starts with the given prefix.The suffix has the same amout of tokens as the prefix and can be made up of more than one sentence. The task is the following: \\
    \\
    \hspace*{0em}According to the title: \{title\}, please complete the following text with a similar amout of words: \{prefix\}\\
    \\
    \hspace*{0em}Return only the completed suffix, without any additional text or explanation.\\
  }%
}
\caption{Prompt used for Probing.}
\label{fig:probingg}
\end{figure}

\subsection{Name Clozae Queries}

\begin{figure}[H]
\noindent\fbox{%
  \parbox{\textwidth}{%
    \hspace*{0em}You have seen the following passage in your training data. What are the proper names that fill in the [MASK] tokens in it? These names are exactly one word long, and are proper names (not a pronoun or any other word). You must make a guess, even if you are uncertain.The [MASK] tokens can be the same or different names, and you should output the names in the order they appear in the passage. Each name should be wrapped in <name> tags. \\
    \\
    \hspace*{0em}Example:\\
    \\
    \hspace*{0em}Input: "The collaboration between [MASK] and [MASK] led to a breakthrough in quantum mechanics."\\
    \hspace*{0em}Output: <name>Einstein</name> <name>Bohr</name>\\
    \\
    \hspace*{0em}Input: "After [MASK] left the room, [MASK] quickly took his place at the table."\\
    \hspace*{0em}Output: <name>John</name> <name>John</name>\\  
    \\
    \hspace*{0em}Input: \{input\}\\
    \hspace*{0em}Output:\\ 
  }%
}
\caption{Prompt used for NCQ where all of the proper names were masked.}
\label{fig:NCQ_mask_all}
\end{figure}

\begin{figure}[H]
\noindent\fbox{%
  \parbox{\textwidth}{%
    \hspace*{0em}You have seen the following passage in your training data. What is the proper name that fills in the [MASK] token in it? This name is exactly one word long, and is a proper name (not a pronoun or any other word). You must make a guess, even if you are uncertain.\\
    \\
    \hspace*{0em}Example:\\
    \\
    \hspace*{0em}Input: "Stay gold, [MASK], stay gold."\\
    \hspace*{0em}Output: <name>Ponyboy</name>\\
    \\
    \hspace*{0em}Input: "The door opened, and [MASK], dressed and hatted, entered with a cup of tea."\\
    \hspace*{0em}Output: <name>Gerty</name>\\  
    \\
    \hspace*{0em}Input: \{input\}\\
    \hspace*{0em}Output:\\ 
  }%
}
\caption{Prompt used for NCQ where one of the proper names was masked.}
\label{fig:NCQ_mask_one}
\end{figure}

\subsection{DE-COP}

\begin{figure}[H]
\noindent\fbox{%
  \parbox{\textwidth}{%
    \hspace*{0em}You are entering a multiple choice questions exam. You should directly answer each question by choosing the correct option. Be concise and straight to the point in your answer. Output only the letter corresponding to the correct answer.\\
    \\
    \hspace*{0em}Question: Which of the following passages is verbatim from the arXiv paper "\{document\_name\}?"\\
  }%
}
\caption{Prompt used for DE-COP where the arXiv based dataset was used.}
\label{fig:DECOP_arxiv}
\end{figure}

\begin{figure}[H]
\noindent\fbox{%
  \parbox{\textwidth}{%
    \hspace*{0em}You are entering a multiple choice questions exam. You should directly answer each question by choosing the correct option. Be concise and straight to the point in your answer. Output only the letter corresponding to the correct answer.\\
    \\
    \hspace*{0em}Question: Which of the following passages is verbatim from the Wikipedia article "\{document\_name\}"?\\
  }%
}
\caption{Prompt used for DE-COP where the Wikipedia based dataset was used.}
\label{fig:DECOP_Wiki}
\end{figure}

\subsection{Familiarity Ranking}

\begin{figure}[H]
\noindent\fbox{%
  \parbox{\textwidth}{%
    \hspace*{0em}Please rank the following three text snippets based on how likely they are to appear in the well-known paper titled: \{title\}. Order them from most likely (1) to least likely (3). Note that the passages are not presented in any particular order.\\
    \\
    \hspace*{0em}Text snippets:\\
    \hspace*{0em}\{chunks\}\\
    \\
    \hspace*{0em}Respond only with a list of integers, for example: 3, 1, 2\\
  }%
}
\caption{Prompt used for FR where the ranking scale from 1 to 3 was used.}
\label{fig:FR_standard}
\end{figure}

\begin{figure}[H]
\noindent\fbox{%
  \parbox{\textwidth}{%
    \hspace*{0em}Please score the following three text snippets based on how likely they are to appear in the well-known paper titled: {title}. Give each of them a score from 10 to 0 on how likley there are to appear in the paper, where 10 is most likely and 0 is least likely.Note that the passages are not presented in any particular order and scores of two chunks can not be the same.\\
    \\
    \hspace*{0em}Text snippets:\\
    \hspace*{0em}\{chunks\}\\
    \\
    \hspace*{0em}Respond only with a list of integers, where each score corresponds to the given chunk, for example: 2, 10, 6\\
  }%
}
\caption{Prompt used for FR where the ranking scale from 0 to 10 was used.}
\label{fig:FR_ranking_3}
\end{figure}

\begin{figure}[H]
\noindent\fbox{%
  \parbox{\textwidth}{%
    \hspace*{0em}Please score the following five text snippets based on how likely they are to appear in the well-known paper titled: {title}. Give each of them a score from 10 to 0 on how likley there are to appear in the paper, where 10 is most likely and 0 is least likely. Note that the passages are not presented in any particular order and scores of two chunks can not be the same.\\
    \\
    \hspace*{0em}Text snippets:\\
    \hspace*{0em}\{chunks\}\\
    \\
    \hspace*{0em}Respond only with a list of integers, where each score corresponds to the given chunk, for example: 2, 10, 6, 9, 8\\
  }%
}
\caption{Prompt used for FR where the ranking scale from 0 to 10 was used for 5 chunks.}
\label{fig:FR_ranking_5}
\end{figure}

\subsection{Paraphrasing}

\begin{figure}[H]
\noindent\fbox{%
  \parbox{\textwidth}{%
    \hspace*{0em}Paraphrase the following sentence into 3 different versions that vary in structure and vocabulary but preserve the meaning. Return only a JSON list of 3 paraphrased strings.\\
    \hspace*{0em}Sentence:\\
  }%
}
\caption{Prompt used for for paraphrasing chunks.}
\label{fig:Paraphrasing}
\end{figure}

\section{Additional Results}
\subsection{Llama}

\begin{table}[H]
\centering
\small
\begin{tabular}{|c|c|c|c|}
\hline Model & TPR & FPR & AUC-ROC  \\
\hline Wikipedia & 0.446 & 0.446 & 0.5 \\
\hline arXiv & 0.541 & 0.427 & 0.557 \\
\hline
\end{tabular}
\caption{Results for the Llama 3.1 70b model with ~20\% null values counted as incorrectly predicted.}
\label{tab:llama20}
\end{table}

\subsection{Highly Likely Members}

\begin{table}[H]
\centering
\small
\begin{tabular}{|l|p{12cm}|}
\hline
\textbf{Model} & \textbf{Rows and Titles} \\
\hline
\multirow{5}{*}{\texttt{gpt\_4o}} 

& Row 2: Rapid Deceleration of Blast Waves Witnessed in Tycho's Supernova Remnant \\
& Row 11: Deep learning for multimessenger core-collapse supernova detection \\
& Row 12: Gravitational and electromagnetic radiation from binary black holes with electric and magnetic charges: Elliptical orbits on a cone \\
& Row 23: The GALAH Survey: Chemical Clocks \\
& Row 26: The DeLLight experiment to observe an optically-induced change of the vacuum index \\
& Row 27: FFCI: A Framework for Interpretable Automatic Evaluation of Summarization \\
& Row 29: A low [CII]/[NII] ratio in the center of a massive galaxy at z=3.7: witnessing the transition to quiescence at high-redshift? \\
& Row 31: Dust evolution: going beyond the empirical \\
& Row 38: An Analysis of Parton Distribution Functions of the Pion and the Kaon with the Maximum Entropy Input \\
& Row 47: Containment strategies after the first wave of COVID-19 using mobility data \\
& Row 48: Political Geography and Representation: A Case Study of Districting in Pennsylvania \\
& Row 52: Domain Generalization via Semi-supervised Meta Learning \\
& Row 53: CRAHCN-O: A Consistent Reduced Atmospheric Hybrid Chemical Network Oxygen Extension... \\
& Row 55: CREDO project \\
& Row 64: Relativistic accretion disk reflection in AGN X-ray spectra at z=0.5--4... \\
& Row 77: CutLang as an Analysis Description Language for Introducing Students to Analyses in Particle Physics \\
& Row 85: Open heavy-flavour production from small to large collision systems with ALICE at the LHC \\
& Row 87: Case Survey Studies in Software Engineering Research \\
& Row 88: On the stress dependence of the elastic tensor \\
& Row 94: The gravitational wave background signal from tidal disruption events \\
& Row 97: General framework for cosmological dark matter bounds using $N$-body simulations \\
\hline
\end{tabular}
\caption{Highly Likely Member Chunks GPT 4o}
\end{table}

\begin{table}[H]
\centering
\small
\begin{tabular}{|l|p{12cm}|}
\hline
\textbf{Model} & \textbf{Rows and Titles} \\
\hline
\multirow{5}{*}{\texttt{gpt3.5\_turbo}}
& Row 14: A Sheaf and Topology Approach to Generating Local Branch Numbers in Digital Images \\
& Row 48: Political Geography and Representation: A Case Study of Districting in Pennsylvania \\
& Row 70: Study of Transverse Spherocity and Azimuthal Anisotropy in Pb-Pb collisions... \\
& Row 76: Exploring British Accents: Modelling the Trap-Bath Split with Functional Data Analysis \\
& Row 80: Gravitational time dilation, free fall, and matter waves \\
\hline
\end{tabular}
\caption{Highly Likely Member Chunks GPT 3.5 Turbo}
\end{table}

\begin{table}[H]
\centering
\small
\begin{tabular}{|l|p{12cm}|}
\hline
\textbf{Model} & \textbf{Rows and Titles} \\
\hline
\multirow{5}{*}{\texttt{gpt4o\_mini}} 
& Row 69: Rate distortion optimization over large scale video corpus with machine learning \\
& Row 77: CutLang as an Analysis Description Language for Introducing Students to Analyses in Particle Physics \\
& Row 90: The Adaptability and Challenges of Autonomous Vehicles to Pedestrians in Urban China \\
\hline
\end{tabular}
\caption{Highly Likely Member Chunks GPT 4o mini}
\end{table}

\begin{table}[H]
\centering
\small
\begin{tabular}{|l|p{12cm}|}
\hline
\textbf{Model} & \textbf{Rows and Titles} \\
\hline
\multirow{5}{*}{\texttt{claude\_3.5\_sonnet}} 
& Row 0: Hierarchical Planning for Resource Allocation in Emergency Response Systems \\
& Row 2: Rapid Deceleration of Blast Waves Witnessed in Tycho's Supernova Remnant \\
& Row 4: Quantitative assessment of the effects of resource optimization and ICU admission policy... \\
& Row 5: Joint super-resolution and synthesis of 1 mm isotropic MP-RAGE volumes... \\
& Row 6: Etching Plastic Searches for Dark Matter \\
& Row 11: Deep learning for multimessenger core-collapse supernova detection \\
& Row 14: A Sheaf and Topology Approach to Generating Local Branch Numbers in Digital Images \\
& Row 18: Characterization of cubic Li$_{2}$$^{100}$MoO$_4$ crystals for the CUPID experiment \\
& Row 19: Neural network representation of electronic structure from $ab$ $initio$ molecular dynamics \\
& Row 20: Progress in the Glauber model at collider energies \\
& Row 21: Gaia DR2 giants in the Galactic dust -- II. Application of the reddening maps and models \\
& Row 23: The GALAH Survey: Chemical Clocks \\
& Row 26: The DeLLight experiment to observe an optically-induced change of the vacuum index \\
& Row 27: FFCI: A Framework for Interpretable Automatic Evaluation of Summarization \\
& Row 28: Modelling Population III stars for semi-numerical simulations \\
& Row 34: Magellanic satellites in $\Lambda$CDM cosmological hydrodynamical simulations of the Local Group \\
& Row 36: Ice Monitoring in Swiss Lakes from Optical Satellites and Webcams using Machine Learning \\
& Row 38: An Analysis of Parton Distribution Functions of the Pion and the Kaon... \\
& Row 40: Precision of The Chinese Space Station Telescope (CSST) Stellar Radial Velocities \\
& Row 42: Gravitational waves from mountains in newly born millisecond magnetars \\
& Row 51: Playing Carcassonne with Monte Carlo Tree Search \\
& Row 52: Domain Generalization via Semi-supervised Meta Learning \\
& Row 56: Multispectral Fusion for Object Detection with Cyclic Fuse-and-Refine Blocks \\
& Row 64: Relativistic accretion disk reflection in AGN X-ray spectra at z=0.5--4... \\
& Row 70: Study of Transverse Spherocity and Azimuthal Anisotropy in Pb-Pb collisions... \\
& Row 71: Lattice strain accommodation and absence of pre-transition phases... \\
& Row 73: DeepFolio: Convolutional Neural Networks for Portfolios with Limit Order Book Data \\
& Row 75: Vortex Dynamics and Phase Diagram in the Electron Doped Cuprate Superconductor... \\
& Row 79: Heating Rates for Protons and Electrons in Polar Coronal Holes... \\
& Row 80: Gravitational time dilation, free fall, and matter waves \\
& Row 82: Role of magnetic exchange interactions in chiral-type Hall effects... \\
& Row 85: Open heavy-flavour production from small to large collision systems with ALICE... \\
& Row 88: On the stress dependence of the elastic tensor \\
& Row 94: The gravitational wave background signal from tidal disruption events \\
& Row 97: General framework for cosmological dark matter bounds using $N$-body simulations \\
\hline
\end{tabular}
\caption{Highly Likely Member Chunks Claude Sonnet 3.5}
\end{table}

\begin{table}[H]
\centering
\small
\begin{tabular}{|l|p{12cm}|}
\hline
\textbf{Model} & \textbf{Rows and Titles} \\
\hline
\multirow{5}{*}{\texttt{llama3.1\_70b}} 
& Row 0: Hierarchical Planning for Resource Allocation in Emergency Response Systems \\
& Row 8: Deep Learning Methods for Screening Pulmonary Tuberculosis Using Chest X-rays \\
& Row 12: Gravitational and electromagnetic radiation from binary black holes... \\
& Row 43: Interior Point Solving for LP-based prediction+optimisation \\
& Row 47: Containment strategies after the first wave of COVID-19 using mobility data \\
& Row 58: A Titan mission using the Direct Fusion Drive \\
& Row 67: The First Months of COVID-19 in Madagascar \\
& Row 69: Rate distortion optimization over large scale video corpus with machine learning \\
& Row 75: Vortex Dynamics and Phase Diagram in the Electron Doped Cuprate Superconductor... \\
& Row 80: Gravitational time dilation, free fall, and matter waves \\
\hline
\end{tabular}
\caption{Highly Likely Member Chunks Llama 3.1 70b}
\end{table}

\end{document}